\documentclass[runningheads]{llncs}
\usepackage{graphicx}
\usepackage{amsmath,amssymb} 
\usepackage{color}

\usepackage{graphicx}
\usepackage{algorithm}
\usepackage{algorithmic}
\newcommand{\etal}{\textit{et al}. }

\newcommand{\eg}{\textit{e}.\textit{g}.}

\usepackage[pagebackref=false,breaklinks=true,letterpaper=true,colorlinks,bookmarks=false]{hyperref}       

\begin{document}

\title{Scene Coordinate Regression with Angle-Based Reprojection Loss for Camera Relocalization} 

\titlerunning{Scene Coordinate Regression with Angle-Based Reprojection Loss}

\authorrunning{X. Li, J. Ylioinas, J. Verbeek and J. Kannala}

\author{Xiaotian Li\inst{1}, Juha Ylioinas\inst{1}, Jakob Verbeek\inst{2} and Juho Kannala\inst{1}}
\institute{Aalto University\\
\email{firstname.lastname@aalto.fi} \and
Univ. Grenoble Alpes, Inria, CNRS, Grenoble INP, LJK, 38000 Grenoble, France\\
\email{firstname.lastname@inria.fr}}

\maketitle

\begin{abstract}
Image-based camera relocalization is an important problem in computer vision and
robotics. Recent works utilize convolutional neural networks (CNNs) to regress for pixels in a query image their corresponding 3D world coordinates in the scene. The final pose is then solved via a RANSAC-based optimization scheme using the predicted coordinates. Usually, the CNN is trained with ground truth scene coordinates, but it has also been shown that the network can discover 3D scene geometry automatically by minimizing single-view reprojection loss. However, due to the deficiencies of the reprojection loss, the network needs to be carefully initialized. In this paper, we present a new angle-based reprojection loss, which resolves the issues of the original reprojection loss. With this new loss function, the network can be trained without careful initialization, and the system achieves more accurate results. The new loss also enables us to utilize available multi-view constraints, which further
improve performance.
\end{abstract}

\section{Introduction}
Image-based camera relocalization is the problem of estimating the 6 DoF camera pose in an environment from a single image. It plays a crucial role in computer vision and robotics, and is the key component for a wide range of applications, such as pedestrian localization and navigation of  autonomous robots \cite{FAB-MAP}, simultaneous localization and mapping (SLAM) \cite{Eade_2006}, and augmented reality \cite{Castle_2008,LynenSBHPS15}.

Many conventional localization methods proposed in the literature \cite{SattlerLK11,SattlerLK12,SattlerLK17} are based on hand-crafted local image features, such as SIFT, ORB, or SURF \cite{Lowe04,RubleeRKB11,Bay06surf}. These methods usually require a 3D point cloud model where each 3D point is associated with the image features from which it was triangulated. Given a query image, its 6D camera pose is recovered by first finding a large set of matches between 2D image features and 3D points in the model via descriptor matching, and then using  a RANSAC-based \cite{fischler_bolles_1981} strategy  to reject outlier matches and estimate the camera pose on inliers. However, these local image features are limited by their hand-crafted feature detectors and descriptors. They are not robust enough for localization in challenging scenarios, and thus limit the use of these conventional methods.

Neural network based localization approaches have been recently explored. For example, PoseNet \cite{kendall2015convolutional} utilizes a convolutional neural network that takes a single RGB image as input and directly regresses the 6 DoF camera pose relative to a scene. Since these methods formulate the camera pose estimation task as a regression problem, which is solved using neural networks, no conventional hand-crafted features are required. These methods can successfully predict camera pose in challenging environments where the conventional methods fail, but their overall accuracy  still lags behind the conventional ones. An alternative neural network based solution is to keep the two-stage pipeline of conventional methods and formulate the first stage of the pipeline for generating 2D-3D correspondences as a learning problem. For example, the recently presented DSAC pipeline \cite{Brachmann_2017_CVPR}  predicts the 6 DoF pose by  first regressing for image patches their 3D positions in the scene coordinate frame, and then determining the camera pose via a RANSAC-based scheme using the produced correspondences. The regression step in the first stage is the so-called scene coordinate regression and the 3D positions are the scene coordinates. Results have shown that these methods achieve state-of-the-art localization accuracy. 

\begin{figure}[t]
\begin{center}
\centering \includegraphics[width=\textwidth]{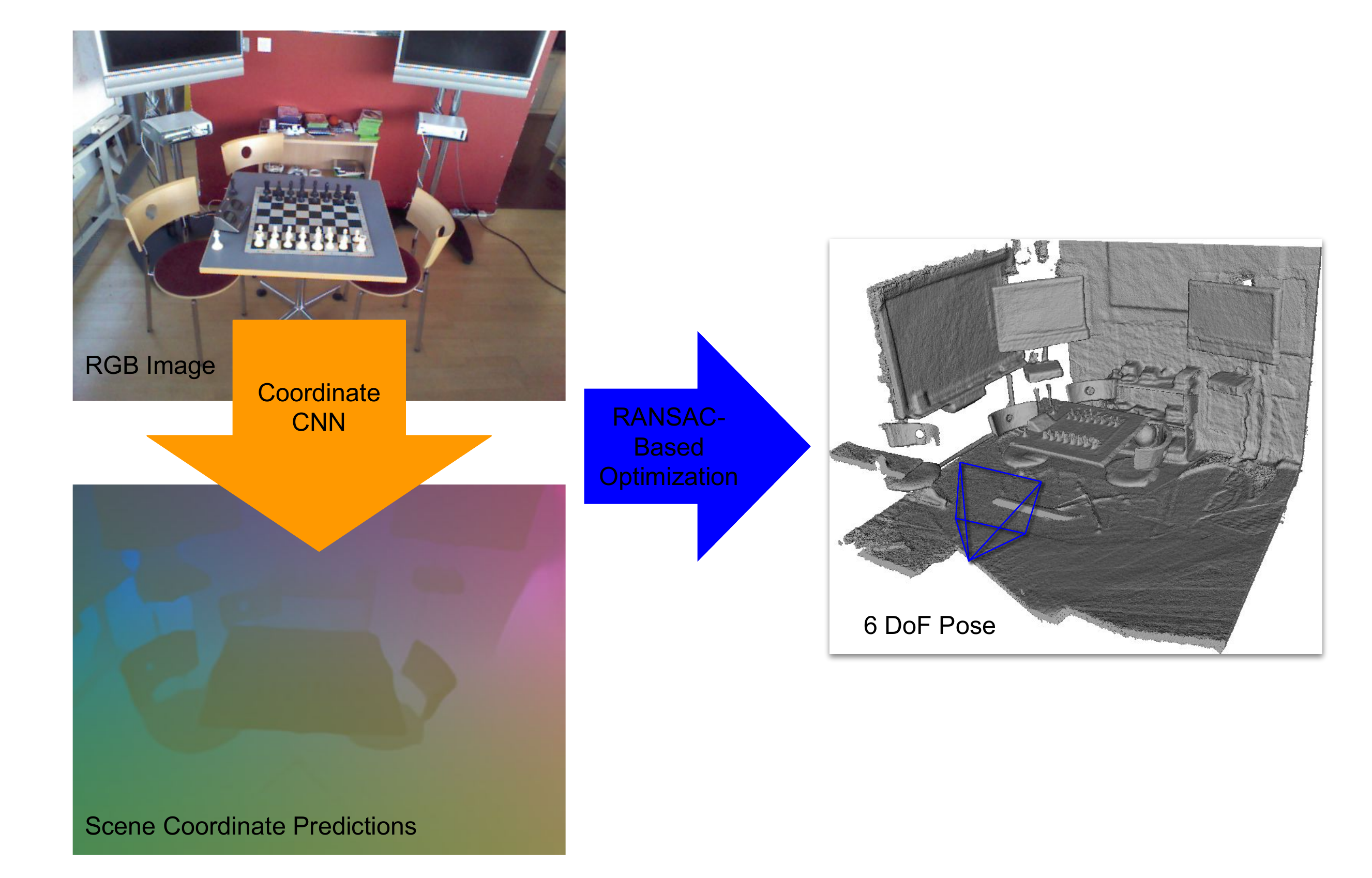}
\end{center}
\caption{Localization pipeline. In this two-stage pipeline, a coordinate CNN first produces scene coordinate predictions from an RGB image, and then the predicted correspondences are fed into a RANSAC-based solver to determine the camera pose. \label{pipeline}} 
\end{figure}

As the improved version of DSAC, DSAC++ \cite{Brachmann_2018_CVPR} has demonstrated that scene coordinate regression can be learned  even without ground truth scene coordinates. A new 3-step training strategy is proposed for training the DSAC++ pipeline. When there are no ground truth scene coordinates available, the first training step is to initialize the network using approximate scene coordinates. In the second training
step, the network is further trained via optimizing reprojection loss which is measured by the sum of residuals between projected 3D points in a 2D image plane. Finally, an end-to-end optimization step is performed. The second step is crucial for the network to discover scene geometry and the first step is also necessary for the second step to work. However, initializing the network using approximate scene coordinates that are far from the ground truth ones might also degenerate localization accuracy.

In order to train the network without an initialization step,  we propose in this work a novel loss function for learning scene coordinate regression. We call this new loss  angle-based reprojection loss. This new loss function has better properties compared to the original reprojection loss, and thus careful initialization is not required. In addition, this new loss allows us to additionally exploit multi-view constraints. 

The contributions of this paper can be summarized as follows:
\begin{itemize}
  \item We present a novel angle-based reprojection loss, which enables the training of coordinate network without careful initialization and improves localization accuracy.
  \item We show that based on this new loss function, we can incorporate multi-view constraints to further improve the accuracy.
\end{itemize}

\section{Related Work}
Image-based camera localization in large-scale environments is often addressed as an image retrieval problem \cite{ZhangK06}. These methods rely on efficient and scalable retrieval approaches to determine the location of a query image. Typically, the query image is matched to a database of geotagged  images and the most similar images to it are retrieved. The location of the query image is then estimated according to the known locations of the retrieved images. These approaches can scale to extremely large urban environments and even entire cites. However, all these methods can only produce an approximate location of the camera and are unable to output an exact  estimate of 6 DoF camera pose.

To directly obtain 6 DoF camera pose with respect to a scene, approaches based on sparse local features such as SIFT \cite{Lowe04} have been proposed. Instead of matching a query image to geotagged database images, these methods usually use a reconstructed 3D point cloud model obtained from Structure-from-Motion \cite{SnavelySS08} to represent the scene.  Each 3D point in the model is associated with one or more local image feature descriptors. Thus, a set of 2D-3D correspondences can be established by matching local image features to 3D points in the model. Based on the 2D-3D correspondences, the camera pose of the query image can be determined by a Perspective-n-Point \cite{KneipSS11} solver inside a RANSAC \cite{fischler_bolles_1981} loop. To scale up these methods to large environments, an efficient and effective descriptor matching step is needed. Therefore, techniques such as prioritized matching \cite{SattlerLK11}, intermediate image retrieval \cite{IrscharaZFB09,Sattler17cvpr}, geometric outlier filtering \cite{ZeislSP15}, and co-visibility filtering \cite{SattlerLK12} have been proposed. However, due to limitations of the hand-crated features, these approaches can fail in challenging scenarios.

PoseNet \cite{kendall2015convolutional} was the first approach to tackle the problem of 6 DoF camera relocalization with deep learning. PoseNet demonstrates the feasibility of directly regressing the 6 DoF camera pose from a query RGB image via a deep CNN. Later works have been proposed to extend this method for better accuracy. For example, in \cite{KendallC15bay}, the authors explore Bayesian Neural Networks to produce relocalization uncertainty of the predicted pose. LSTM-Pose \cite{Walch_2017_ICCV} utilizes LSTM units \cite{Hochreiter1997} on the output of CNN to extract better features for localization. Another variant, Hourglass-Pose \cite{MelekhovYKR17}, makes use of the encoder-decoder hourglass architecture which can  preserve the fine-grained information of input image. Moreover, \cite{Kendall_2017_CVPR} demonstrates that PoseNet performance can be improved by leveraging geometric loss functions. However, these methods are still outperformed by conventional sparse feature based methods. More recently, two multitask models VlocNet \cite{vlocnet} and VlocNet++ \cite{vlocnet++}   have been introduced. These models operate on consecutive monocular images and utilize auxiliary learning during training. Remarkably, they can offer promising localization performance that surpasses the sparse feature based methods.

Unlike PoseNet, Laskar \etal introduce a deep learning based localization framework, which relies on image retrieval and relative camera pose estimation \cite{LaskarMKK17}. This method  requires no scene-specific training and can generalize well to previously unseen scenes. Similarly, Taira \etal put forth an image retrieval based localization system for large-scale indoor environments \cite{inloc}. After retrieval of candidate poses, the pose is estimated based on dense matching. A final pose verification step via virtual view synthesis can further improve the robustness of the system. To achieve robust visual localization under a wide range of viewing conditions, semantic information has also been exploited in \cite{semanticloc}.

Our work is most closely related to methods based on the scene coordinate regression framework. The original scene coordinate regression pipeline is proposed for RGB-D camera relocalization \cite{SCoRF}. This method formulates descriptor matching as a regression problem and applies a  regression forest \cite{Criminisi_2013_DFC} to produce 2D-3D correspondences from an RGB-D input image. Similarly to a sparse feature based  pipeline, the final camera pose is recovered from the correspondences via a RANSAC-based solver. Since generating correspondences is directly achieved by the regression forest, no traditional feature extraction, feature description, or feature matching processes are required. This method has been further extended in later works \cite{Guzman-RiveraKGSSFI14,ValentinNSFIT15,CavallariGLVST17}. Currently, practical low-cost devices are usually equipped with  RGB cameras only. However, these methods still require a depth signal at both training and test time.

To localize RGB-only images,  an auto-context random forest is adopted in the pipeline in \cite{BrachmannMKYGR16}. In \cite{Massiceti_icra}, Massiceti \etal explored random-forest-to-neural-network mapping strategy. Recently, Brachmann et al. proposed a differentiable version of RANSAC (DSAC) \cite{Brachmann_2017_CVPR} and presented a localization system based on it. In the DSAC pipeline, two CNNs are adopted for predicting scene coordinates and for scoring hypotheses respectively. Since the entire pipeline is differentiable, an end-to-end optimization step can be performed using ground truth camera poses. In contrast to DSAC, which adopts a patch-based network for scene coordination regression, a full-frame network considering global image appearance is presented in \cite{Li2018} and a data augmentation strategy is proposed to ensure the accuracy. However, all of these methods require ground truth scene coordinates for training.  DSAC++ \cite{Brachmann_2018_CVPR}, the successor version of DSAC, demonstrates the feasibility of learning scene coordinate regression without scene coordinate ground truth. This is achieved by first initializing the predictions with approximate scene coordinates and then optimizing reprojection errors. This method is the current state-of-the-art on the 7-Scenes dataset \cite{SCoRF} and the Cambridge Landmarks dataset \cite{kendall2015convolutional}. 

\section{Method}

In this work, we follow the two-stage pipeline of DSAC and DSAC++ for RGB-only camera relocalization. In the first stage, given an RGB image, a coordinate CNN is adopted to generate dense 3D scene coordinate predictions to form 2D-3D correspondences. In the second stage, a RANSAC-based scheme is performed to generate pose hypotheses with their scores and  determine the final pose estimate which can be further refined. The overall pipeline is illustrated in Figure \ref{pipeline}.

DSAC first demonstrates how to incorporate CNNs into the two-stage pipeline and achieves state-of-the-art results. DSAC++ further addresses the main shortcomings of the DSAC pipeline and achieves substantially more accurate results. More importantly, unlike previous works which require scene coordinate ground truth generated using RGB-D training images or available 3D model during training, DSAC++ is able to discover 3D scene geometry automatically. That is, the coordinate CNN can be trained from RGB images only with their ground truth camera poses, and no ground truth scene coordinates or depth information is needed. This is achieved by first using  approximate scene coordinates to initialize the network and then optimizing reprojection loss, which is calculated using the ground truth poses. In the initialization step, the approximate scene coordinates are generated to have a constant distance $d$ from the camera plane and serve as the dummy ground truth for training. The reprojection loss can then help the network to recover the scene geometry, but it may not work without the initialization step, even though the initialization step itself provides poor localization performance. However, the DSAC++ pipeline still has a main drawback. That is, a proper value for constant distance $d$ should be selected carefully, since a value that is far off the actual range of distances can result in poor accuracy \cite{Brachmann_2018_CVPR}. 

In the following, we present a new angle-based reprojection loss that allows us to train the coordinate CNN without careful initialization. We explain the deficiencies of the original reprojection loss and present our new loss function in Section \ref{sec:3.1}. This new loss function also enables us to utilize available multi-view reprojection and photo-consistency constraints, which we discuss in Section \ref{sec:3.2} and Section \ref{sec:3.3} respectively.
\subsection{Angle-Based Reprojection Loss}\label{sec:3.1}
For a training image $I_i$, the reprojection loss is calculated using the scene coordinate predictions and the ground truth pose. It can be written as\footnote{We omit the conversions between homogeneous coordinates and Cartesian coordinates for simplicity. This also applies to equations in the rest of the paper.}:
\begin{equation}
\label{reproloss}
\mathcal{L}_{rep} = 	  \sum_{k\in P_i}\| Ch^{-1}_iy_k(I_i;w)-p_k^i\| 
\end{equation}
where $C$ is the intrinsic matrix, $h_i$ is the camera pose for image $I_i$, $P_i$ is the set of all points in image $I_i$, $y_k(I_i;w)$ is the scene coordinate prediction for point $k$ with input image $I_i$ and learnable parameters $w$, $p_k^i$ is the 2D position of point $k$ in image $I_i$. 

As mentioned above, the DSAC++ pipeline requires a carefully selected distance $d$ for generating approximate scene coordinates when training without ground truth scene coordinates. However, improper selection of $d$ may also lead to poor performance. Therefore, one may wonder if it is possible to train the network without an initialization step. Unfortunately,  if we train the network directly using the reprojection loss without the initialization step, the training might not converge due to unstable gradients. The unstable gradients are caused by minimizing the reprojection loss at the beginning of training when the predictions of the network could be behind the camera or close to the camera center.

\begin{figure}
	\centering \includegraphics[height=7cm]{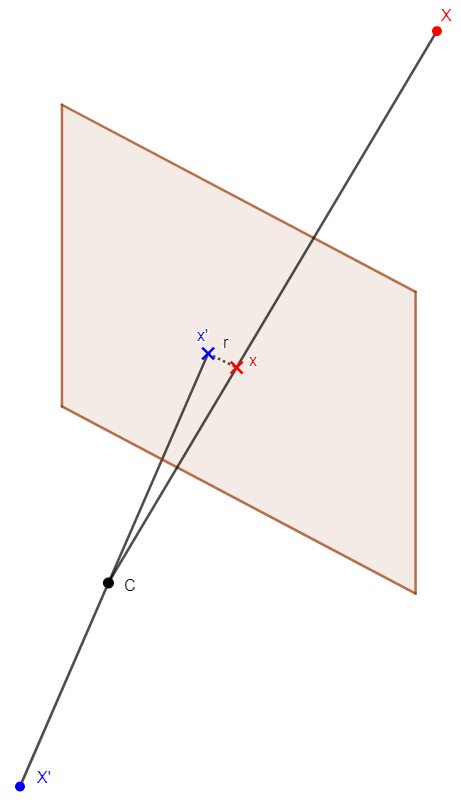}
~
\centering \includegraphics[height=7cm]{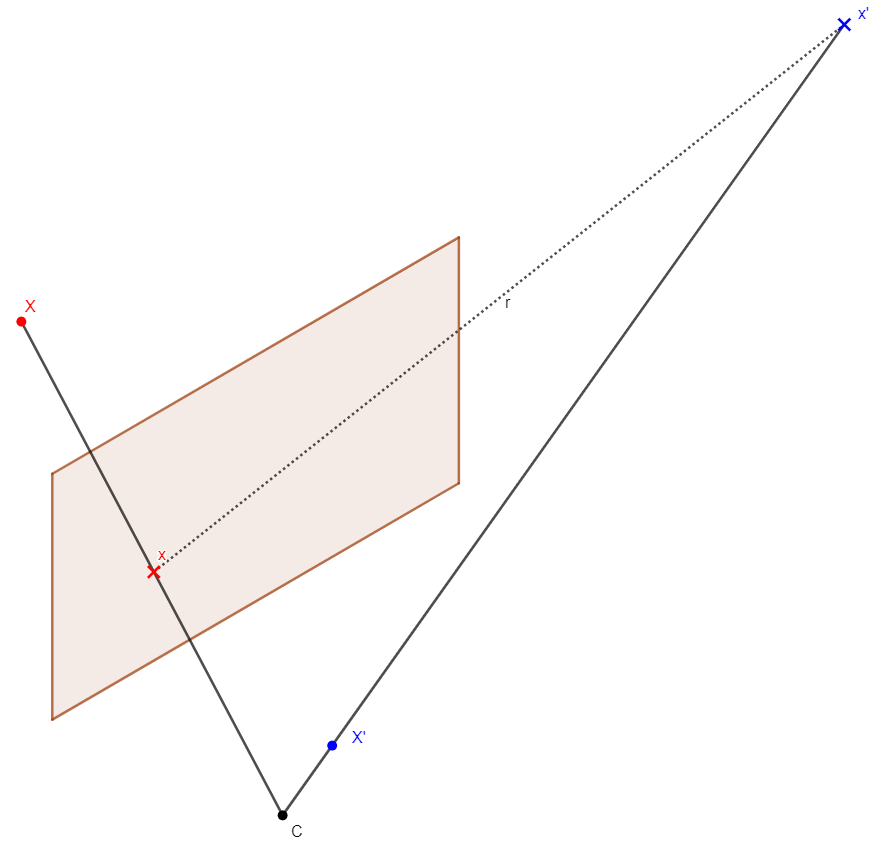}
~
\centering \includegraphics[height=10cm]{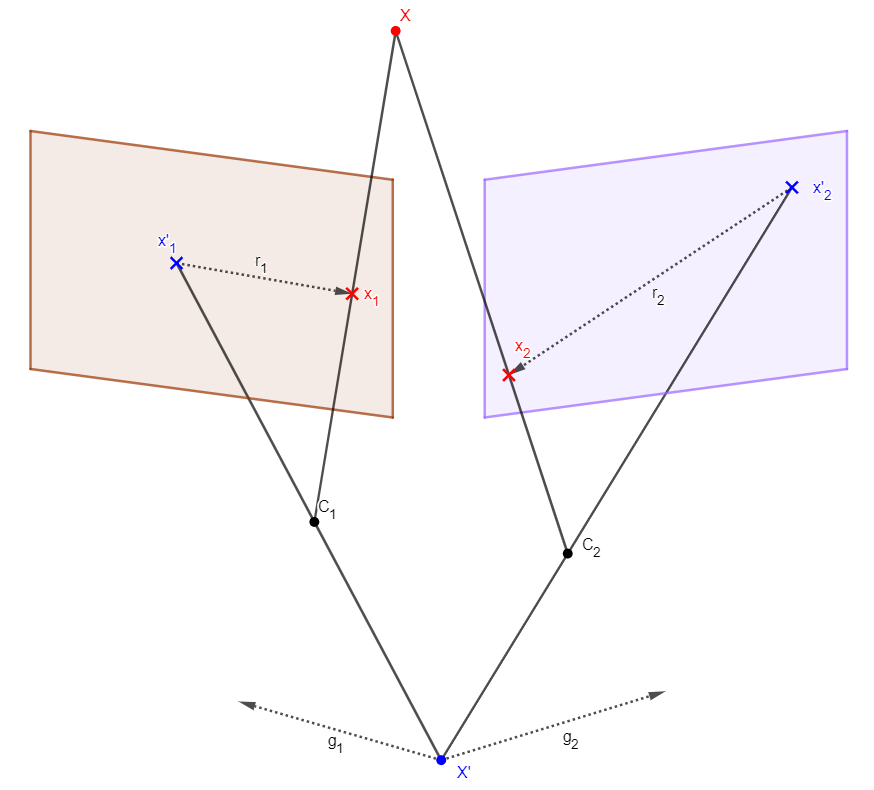}
	\caption{Top left: when predictions are behind the camera, loss can still be 0. Top right: reprojected predictions can be easily far from the ground truth. Bottom: predictions behind the camera can cause inconsistent gradients. \label{loss_1}} 
\end{figure}

The shortcomings of the reprojection loss are illustrated in Figure \ref{loss_1}. As we can see, the reprojection loss does not constrain the predictions to be in front of the camera, and the loss could be 0 even when the predictions are behind the camera. Therefore, in such cases, the reprojection loss cannot help the network to discover the true geometry of the scene. In addition, when the $z$ coordinate of a prediction in the camera coordinate frame is close to zero, the corresponding projected point in the image plane could be extremely far away from the ground truth one, resulting in extremely large loss value and exploding gradients. More importantly, we believe the ability of the network to discover the scene geometry automatically comes from the patch-based nature of the predictions. The output neurons of the fully-convolutional network used in the DSAC++ pipeline have a limited receptive field. Since local patch appearance is relatively invariant w.r.t. to viewpoint change, the explicitly applied single-view reprojection loss could be considered as implicit multi-view constraints. However, when the predictions are behind camera, minimizing reprojection loss to fulfill multi-view constraints could lead to inconsistent gradients. This again prevents the network from discovering the true geometry of a scene.

The aforementioned deficiencies restrict the use of the reprojection loss. In particular,  before using the reprojection loss, we need to initialize the predictions to be in front of  the camera using a constant distance $d$. However, this might lead to overfitting to the dummy ground truth scene coordinates, and thus affect the localization accuracy.

\begin{figure}
	\centering \includegraphics[height=8cm]{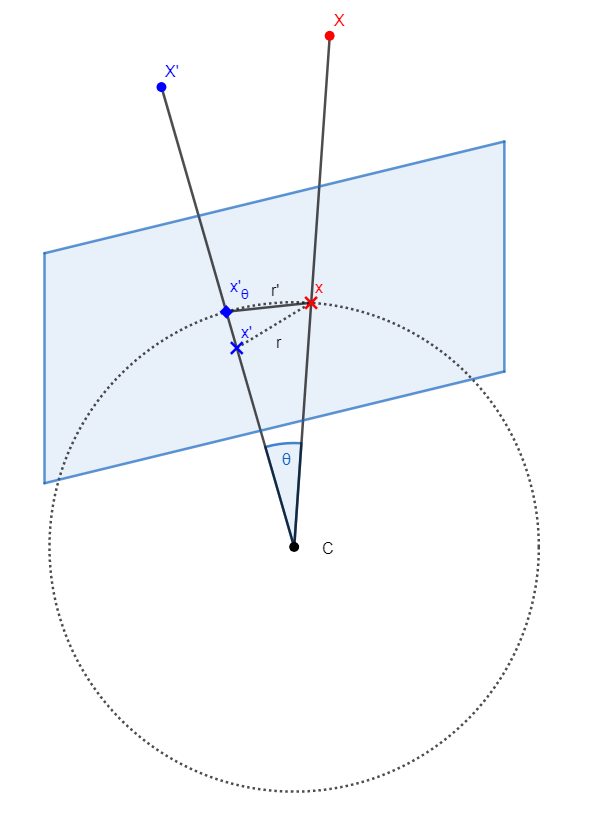}
~
\centering \includegraphics[height=9cm]{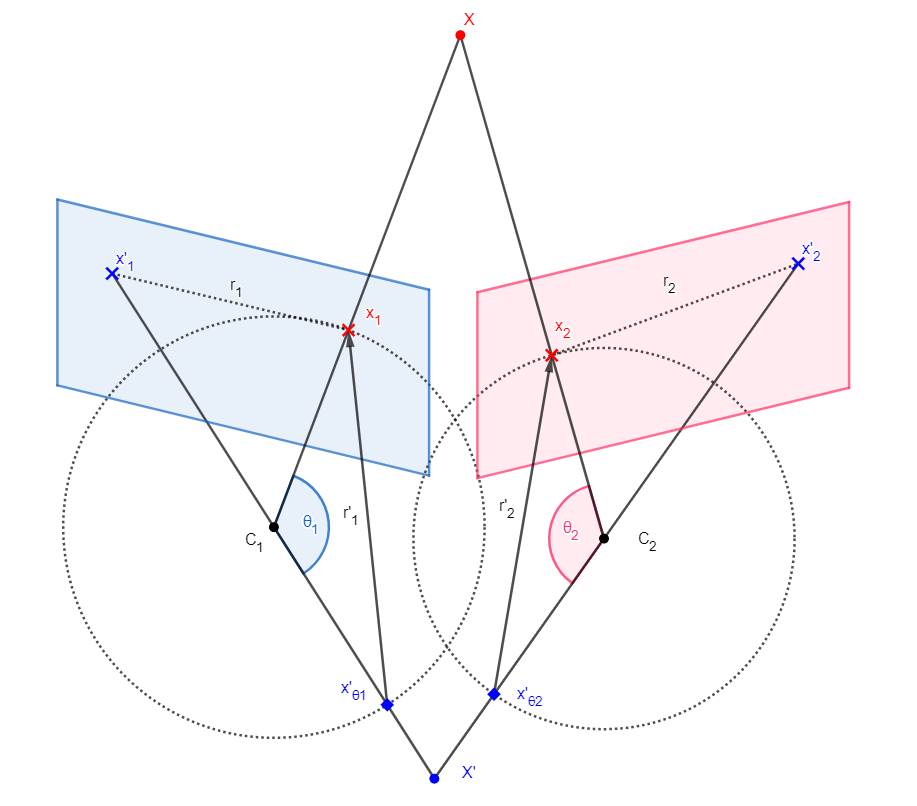}

	\caption{Top: to minimize $\theta$, we minimize $r'$. Bottom: the new loss enforces predictions to be in front of the camera and causes no unstable gradients. \label{loss_2}} 
\end{figure}

To address these issues, we present a new angle-based reprojection loss. Instead of minimizing the distance between the projected ground truth and projected prediction in the image plane, we seek to minimize the angle $\theta$ between two rays that share the camera center as the common endpoint and pass through the scene coordinate prediction and ground truth respectively. To be specific, we minimize the distance between two points in which the two rays intersect a sphere with center at the camera center and radius equal to the distance between the projected ground truth and the camera center, as shown in Figure \ref{loss_2}. More formally, for a training image $I_i$, the angle-based reprojection loss is defined as\footnote{We assume $f_x=f_y=f$ in this paper, but the loss function is also applicable when $f_x\neq f_y$.}:
\begin{equation}
\label{angreproloss}
\mathcal{L}_{ang} = 	  \sum_{k\in P_i}\| \frac{\|d_k^i\|}{\|D_k^i\|} h^{-1}_iy_k(I_i;w)-fC^{-1}p_k^i\| 
\end{equation}
where $f$ is the focal length, $d_k^i=fC^{-1}p_k^i$, and $D_k^i=h^{-1}_iy_k(I_i;w)$. 

As we can see, the new angle-based reprojection loss forces the predictions to be in front of camera, since the loss function has large value when predictions are behind camera. Moreover, it no longer causes unstable gradients when predictions are behind camera or  the $z$ coordinates of the predictions in the camera frame are close to zero. In addition, it is easy to see that the new loss function is approximately equal to the original one when predictions are close to ground truth scene coordinates. Assume $h^{-1}_iy_k(I_i;w)=(X_k^i,Y_k^i,Z_k^i)^T$ , $p_k^i = (x_k^i,y_k^i)^T$, and the ground truth $\dot{D_k^i}=(\dot{X}_k^i,\dot{Y}_k^i,\dot{Z}_k^i)^T$. When predictions are close to ground truth, we have $\frac{\|d_k^i\|}{\|D_k^i\|}\approx\frac{\|d_k^i\|}{\|\dot{D}_k^i\|}=\frac{f}{\dot{Z}_k^i}\approx\frac{f}{Z_k^i}$. Therefore: 
\begin{equation}
\begin{aligned}
\label{appro}
\mathcal{L}_{ang} &= 	  \sum_{k\in P_i}\| \frac{\|d_k^i\|}{\|D_k^i\|} h^{-1}_iy_k(I_i;w)-fC^{-1}p_k^i\| \\
&= 	  \sum_{k\in P_i}\sqrt{(\frac{\|d_k^i\|}{\|D_k^i\|}X_k^i+c_x-x_k^i)^2+(\frac{\|d_k^i\|}{\|D_k^i\|}Y_k^i+c_y-y_k^i)^2+(\frac{\|d_k^i\|}{\|D_k^i\|}Z_k^i-f)^2}  \\
&\approx 	  \sum_{k\in P_i}\sqrt{(\frac{f}{Z_k^i}X_k^i+c_x-x_k^i)^2+(\frac{f}{Z_k^i}Y_k^i+c_y-y_k^i)^2} \\
&= \sum_{k\in P_i}\| Ch^{-1}_iy_k(I_i;w)-p_k^i\| = \mathcal{L}_{rep}
\end{aligned}
\end{equation}
where $c_x$ and $c_y$ are parameters for principal point offset. This is the reason that we call the new loss function angle-based reprojection loss.

Note that in order to minimize the angle $\theta$, one may also maximize the dot product of the two unit vectors of the two rays, but the resulting loss function no longer has the property to be approximately equal to the reprojection loss when predictions are close to ground truth scene coordinates.

\subsection{Multi-View Reprojection Loss}\label{sec:3.2}
Although with only single-view reprojection loss, the network can already approximately recover the scene geometry, sometimes it is still difficult for the network to learn accurate scene coordinate predictions, \eg, when ground truth poses are erroneous or there are ambiguities in the scene. Therefore, if we could exploit explicitly multi-view constraints during training, we might achieve more accurate localization performance.

One way to incorporate multi-view constraints is to utilize available multi-view correspondences to form multi-view reprojection loss. The correspondences might come from a reconstructed 3D model or can be directly generated by correspondence matching. 

Since the original reprojection loss is problematic when predictions are behind camera, it is more likely to cause problem in the multi-view cases. Therefore, we formulate the multi-view reprojection loss based on our new angle-based reprojection loss. For one training image, the loss function can be written as: 
\begin{equation}
\label{multi}
\begin{aligned}
\mathcal{L}_{multi} &= 	 \sum_{k\in P_i^-}\| \frac{\|d_k^i\|}{\|D_k^i\|} h^{-1}_iy_k(I_i;w)-fC^{-1}p_k^i\|\\
&+\lambda\sum_{k\in P_i^*}\sum_{j\in \mathcal{I}_{k}\cup\lbrace I_i \rbrace}\| \frac{\|d_k^j\|}{\|D_k^j\|}h^{-1}_jy_k(I_i;w)-fC^{-1}p_k^j\|
\end{aligned}
\end{equation}
where $\lambda$ is the balance weight, $P_i^-$ is the set of all points in image $I_i$ which have no corresponding points, $P_i^*$ is the set of points in image $I_i$ which have corresponding points in other training images, $\mathcal{I}_{k}$ is the set of images in which point $k$ has a corresponding point. As we can see, $ \sum_{k\in P_i^-}\| \frac{\|d_k^i\|}{\|D_k^i\|} h^{-1}_iy_k(I_i;w)-fC^{-1}p_k^i\|$ encodes single-view constraints, and $\sum_{k\in P_i^*}\sum_{j\in \mathcal{I}_{k}\cup\lbrace I_i \rbrace}\| \frac{\|d_k^j\|}{\|D_k^j\|}h^{-1}_jy_k(I_i;w)-fC^{-1}p_k^j\|$ encodes multi-view constraints.
\subsection{Photometric Reconstruction Loss}\label{sec:3.3}
If training data consists of sequences of images, it is also possible to constrain the scene coordinate predictions using photometric reconstruction loss. Photometric reconstruction loss has recently become a dominant strategy for solving many geometry related learning problems. It enables neural networks to learn many tasks, such as monocular depth \cite{GargBR16,GodardAB16}, ego-motion \cite{zhou2017unsupervised}, and optical flow \cite{YuHD16}, in a self-supervised manner.

For learning scene coordinate regression, given a training image $I_i$ and a neighboring image $I_j$, we can reconstruct the training image by sampling pixels from the neighboring image based on the scene coordinate predictions and the pose of the neighboring image. Specifically, for a point $k$ in the training image $I_i$, we can project it to the neighboring image and obtain its 2D coordinates in the neighboring image:  
\begin{equation}
\label{2dcoord}
p_k^j = 	   Ch^{-1}_jy_k(I_i;w) 
\end{equation}
Then, the pixel value for point $k$ in the reconstructed image $\hat{I}_i$ can be sampled using bilinear sampler based on the coordinates  $p_k^j$. That is, the sampled pixel value is the weighted sum of the values of the four neighboring pixels of $p_k^j$.

Following \cite{GodardAB16}, our photometric reconstruction loss is a combination of an $L1$ loss term and a simplified SSIM loss term with a $3\times3$ block filter:
\begin{equation}
\label{pr}
\mathcal{L}_{pr} = 	  \sum_{k\in P_i}(1-\alpha)\mid \hat{I}_i^k - I_i^k\mid +\alpha\frac{1-SSIM(\hat{I}_i^k,I_i^k)}{2}
\end{equation}
where  $\alpha=0.85$.

Since in the beginning of training, the scene coordinate predictions could be quite far from the ground truth ones, the projected 2D coordinates of points in the neighboring image might also be far from valid pixels. In other words, we are unable to obtain valid gradients by minimizing the photometric reconstruction loss. Therefore, using only photometric reconstruction loss for learning scene coordinate regression may not work. Thus, we pair it with our angle-base reprojection loss, and the final loss function can be written as:
\begin{equation}
\label{angpr}
\mathcal{L}_{ang+pr} = 	  \mathcal{L}_{ang} + \lambda\mathcal{L}_{pr}
\end{equation}
\section{Experiments}
We evaluate the performance of our approach on the 7-Scenes dataset \cite{SCoRF} and the Cambridge Landmarks dataset \cite{kendall2015convolutional}, which are described in Section \ref{sec:4.1}. The implementation details are explained in Section \ref{sec:4.2} and the  results are given in Section \ref{sec:4.3}.

\subsection{Evaluation Dataset}\label{sec:4.1}
The 7-Scenes dataset \cite{SCoRF} is a widely used RGB-D dataset  provided by Microsoft Research. It consists of seven different indoor environments, and exhibits several challenges, such as motion blur, illumination changes, textureless surfaces, repeated structures, reflections, and sensor noise. Each scene contains multiple sequences of tracked RGB-D camera frames and these sequences are split into training and testing data. The RGB-D images are captured using a handheld Kinect camera at $640\times480$ resolution and associated with 6 DoF ground truth camera poses obtained via the KinectFusion system \cite{IzadiKHMNKSHFDF11,NewcombeIHMKDKSHF11}. 

The Cambridge Landmarks \cite{kendall2015convolutional} is an outdoor RGB relocalization dataset, which consists of six scenes from around Cambridge University. Each scene is composed of hundreds or thousands of frames, which are divided into training and test sequences.  The 6 DoF ground truth camera poses of the images are generated with structure from motion techniques \cite{Wu13}.

In contrast to some previous works, ground truth scene coordinates are not necessary for learning scene coordinate regression in this work. Therefore, we use only RGB images at both training and test time.

\subsection{Implementation Details}\label{sec:4.2}
Note that the following implementation details are for the experiments on the 7-Scenes dataset. For the experiments on the Cambridge Landmarks dataset, we simply follow DSAC++ \cite{Brachmann_2018_CVPR} except that we train the network  with our angle-based reprojection loss only for $450K$ iterations,  we halve the learning rate every $50K$ iterations after $250K$ iterations, and we perform no gradient clipping.

We adopt the FCN network architecture used in the DSAC++ pipeline which takes
a $640\times480$ image as input and produces $80 \times 60$ scene coordinate predictions. For implementation simplicity, unlike DSAC++, in the second stage of the pipeline, we use a non-differentiable version of the RANSAC-based scheme for pose estimation, which scores hypotheses using inlier count instead of soft inlier count. The details of the used pose estimation algorithm are explained in \cite{Li2018}. In this paper, we mainly focus on the loss functions for learning scene coordinate regression in the first stage of the pipeline, and thus the entire pipeline is not required to be end-to-end trainable. However, it is also straightforward to adopt the differentiable optimization strategy and perform an extra end-to-end optimization step at the end of the training as described in \cite{Brachmann_2018_CVPR}.

We train the network from scratch without the scene coordinate initialization training step for $600K$ iterations with a batch size of 1 using the Adam \cite{Adam} optimizer where $\beta_1=0.9$, $\beta_2=0.999$, and $\epsilon = 10^{-8}$. The initial learning rate is set to $0.0001$ and is halved after $180K$, $360K$, $420K$, $480K$ and $540K$ iterations.  Following \cite{Brachmann_2018_CVPR}, training images are randomly shifted by a maximum of 8 pixels, horizontally and vertically, to make full use of the training data, since the network produces one prediction for each $8\times8$ image block.

When training the network using the multi-view reprojection loss, for efficiency, we do not use all the correspondences to calculate the latter term of the loss. In particular, we calculate $\sum_{k\in P_i^*}\sum_{j\in \lbrace I_i, I_m\rbrace}\| \frac{\|d_k^j\|}{\|D_k^j\|}h^{-1}_jy_k(I_i;w)-fC^{-1}p_k^j\|$
instead of $\sum_{k\in P_i^*}\sum_{j\in \mathcal{I}_{k}\cup\lbrace I_i \rbrace}\| \frac{\|d_k^j\|}{\|D_k^j\|}h^{-1}_jy_k(I_i;w)-fC^{-1}p_k^j\|$ where $I_m$ is an image randomly selected from $\mathcal{I}_{k}$. The co-visibility information used to form the multi-view constraints is extracted from a sparse model constructed with Colmap \cite{SchonbergerF16}. Moreover,  we set the balance weight $\lambda=60$. 

When computing the photometric reprojection loss, the neighboring image is randomly selected from the same sequence with difference in index less or equal to 10. Since the network produces $80 \times 60$ scene coordinate predictions, we resize both the training images (after shifting) and the neighboring images to $80 \times 60$, and the camera intrinsic parameters are also adjusted accordingly.  The balance weight $\lambda$ is set to 20.

\subsection{Results}\label{sec:4.3}
The experimental results of our method on the  7-Scenes dataset and the Cambridge Landmarks dataset are given in Table \ref{re_all}. For the Cambridge Landmarks dataset, we report median localization errors of our method trained with single-view constraints only. For the 7-Scenes dataset, we report both median localization errors and accuracy measured as the percentage of query images for which the camera pose error is below $5^{\circ}$ and 5cm, as the latter one can better represent the localization performance. Our method is compared to the state-of-the-art DSAC++ method trained without an accurate 3D scene model.

\begin{table}
\caption{Results of our method on the 7-Scenes dataset and the Cambridge Landmarks dataset. We compare the performance of our method to DSAC++. Note that Complete denotes the accuracy measured by the $5^{\circ}$, 5cm criterion  on the combined set of the test frames (17000) of all scenes for the 7-Scenes dataset. \label{re_all}}
\begin{center}
\resizebox{\textwidth}{!}{%
\begin{tabular}{c|c|c|c|c|}
\cline{2-5}
                                  & \multicolumn{4}{c|}{$5^{\circ}$, 5cm (Median Error)}                                                                    \\ \hline
\multicolumn{1}{|c|}{Scene}       & DSAC++ \cite{Brachmann_2018_CVPR}                      & Ours                        & Ours+multi-view              & Ours+photometric            \\ \hline
\multicolumn{1}{|c|}{Chess}       & 93.8\% (2cm, $0.7^{\circ}$)  & 95.1\% (2cm, $0.6^{\circ}$) & 96.1\% (2cm, $0.6^{\circ}$)  & 96.0\% (2cm, $0.6^{\circ}$) \\ 
\multicolumn{1}{|c|}{Fire}        & 75.6\% (3cm, $1.1^{\circ}$)  & 84.0\% (3cm, $1.0^{\circ}$) & 88.6\% (2cm, $1.0^{\circ}$)  & 86.4\% (2cm, $1.0^{\circ}$) \\ 
\multicolumn{1}{|c|}{Heads}       & 18.4\% (12cm, $6.7^{\circ}$) & 80.5\% (2cm, $1.1^{\circ}$) & 86.9\% (2cm, $1.0^{\circ}$)  & 83.2\% (2cm, $1.1^{\circ}$) \\ 
\multicolumn{1}{|c|}{Office}      & 75.4\% (3cm, $0.8^{\circ}$)  & 80.4\% (3cm, $0.8^{\circ}$) & 80.6\% (3cm, $0.8^{\circ}$)  & 81.6\% (3cm, $0.8^{\circ}$) \\ 
\multicolumn{1}{|c|}{Pumpkin}     & 55.9\% (5cm, $1.1^{\circ}$)  & 56.8\% (4cm, $1.1^{\circ}$) & 60.3\% (4cm, $1.1^{\circ}$)  & 59.2\% (4cm, $1.1^{\circ}$) \\ 
\multicolumn{1}{|c|}{Red Kitchen} & 50.7\% (5cm, $1.3^{\circ}$)  & 59.9\% (4cm, $1.2^{\circ}$) & 61.9\% (4cm, $1.2^{\circ}$)  & 60.0\% (4cm, $1.2^{\circ}$) \\
\multicolumn{1}{|c|}{Stairs}      & 2.0\% (29cm, $5.1^{\circ}$)  & 2.9\% (25cm, $4.5^{\circ}$) & 11.3\% (13cm, $3.4^{\circ}$) & 4.7\% (22cm, $4.9^{\circ}$) \\ \hline
\multicolumn{1}{|c|}{Complete}    & 60.4\%                      & 69.2\%                     & 71.8\%                       & 70.4\%                      \\ \hline
\multicolumn{1}{|c|}{Great Court}      &  (66cm, $0.4^{\circ}$)  &  (51cm, $0.3^{\circ}$) & -  & - \\
\multicolumn{1}{|c|}{K. College}      &  (23cm, $0.4^{\circ}$)  &  (18cm, $0.3^{\circ}$) & -  & - \\
\multicolumn{1}{|c|}{Old Hospital}      &  (24cm, $0.5^{\circ}$)  &  (19cm, $0.4^{\circ}$) & -  & - \\
\multicolumn{1}{|c|}{Shop Facade}      &  (9cm, $0.4^{\circ}$)  &  (7cm, $0.3^{\circ}$) & -  & - \\
\multicolumn{1}{|c|}{St M. Church}      &  (20cm, $0.7^{\circ}$)  &  (25cm, $0.7^{\circ}$) & -  & -
\\ \hline
\end{tabular}}
\end{center}
\end{table}

According to the results, for the 7-Scenes dataset, when training the network using our new angle-based reprojection loss without initializing the network predictions with constant distance value, the fraction of accurately localized test images is improved by 8.8 percentage points compared to DSAC++. Similarly, for the Cambridge Landmarks dataset, our method achieves better median pose accuracy on four out of five scenes. Moreover, for the 7-Scenes dataset, the accuracy can be further improved by utilizing either multi-view correspondences or  photometric reconstruction metric. Compared to using single-view constraints only, we observed that the additional loss terms could make the network produce more accurate scene coordinate predictions during training. This means that incorporating multi-view constraints could help the network better discover scene geometry, and thus leads to improved localization performance. 

We also attempted to train the network using the original reprojection loss without the scene coordinate initialization step. However, we found that for most of the scenes, training could not converge in both single-view and multi-view cases. In addition, when using the photometric reconstruction loss, for all the scenes,  we observed that the training of the network without the angle-based reprojection loss term  would always get stuck in a local minimum with large scene coordinate prediction error, resulting in a completely failed localization system. 

When training with either the multi-view reprojection loss or the photometric reconstruction loss, the balance weight $\lambda$ is important for achieving good localization results. If the weight is set too small, the multi-view constraints will have no effect on  training. However, if it is set too large, we found that the localization accuracy could also drop. For example, when training with the photometric reconstruction loss with $\lambda = 50$, test accuracy for Office decreases to 79.5\%. It would be interesting to explore learnable weighting strategy presented in \cite{Kendall_2017_CVPR} in future work. 

Note that our method is still less accurate compared to the DSAC++ method trained with a 3D model (76.1\%), as our method is trained without access to adequate scene geometry information. When training with the multi-view reprojection loss,  only a sparse set of points have corresponding points in other images (typically about 1000 points for a $640\times480$ image). That is, the multi-view constraints are active for only a small portion of the points in a training image. We believe that with denser correspondence information, our method could achieve better results. For the photometric reconstruction loss, adding an additional loss term similar to the left-right consistency proposed in \cite{GodardAB16}, which enforces predictions to be consistent between different views,  might be also helpful to further improve the accuracy, but we did not explore it.

\section{Conclusion}

In this work, we have presented a new angle-based reprojection loss for learning scene coordinate regression for image-based camera relocalization. Our novel loss function makes it possible to train the coordinate CNN without a scene coordinate initialization step, resulting in improved localization accuracy. Moreover, this novel loss function allows us to explore available multi-view constraints, which can further improve performance. 

\section*{Acknowledgements}
Authors acknowledge funding from the Academy of Finland
(grant numbers 277685, 309902). This work has also been partially supported by the grant “Deep in France” (ANR16-CE23-0006) and LabEx PERSYVAL (ANR-11-LABX0025-01).

\bibliographystyle{splncs04}
\bibliography{bib}
\end{document}